\documentclass[letterpaper, 10 pt, conference]{ieeeconf}  

\IEEEoverridecommandlockouts                              

\usepackage{graphics} 
\graphicspath{ {./images/} }
\usepackage{epsfig} 
\usepackage[ruled,vlined,linesnumbered]{algorithm2e}
\usepackage{algorithmic}

\title{\LARGE \bf
ARC Nav - A 3D Navigation Stack for Autonomous Robots
}

\author{Vishwas N.S$^{1*}$, Srikrishna B.R$^{2*}$ and Sudarshan T.S.B$^{3}$

\thanks{* Equal Contribution}%
\thanks{$^{1}$Vishwas N.S is a student at Department of Computer Science and Engineering,
        PES University, Bangalore, India.
        {\tt\small vishwas3151999@gmail.com}}%
\thanks{$^{2}$Srikrishna B.R is a student at Department of Electrical and Electronics Engineering,
        PES University, Bangalore, India.
        {\tt\small brsrikrishna@gmail.com}}%
\thanks{$^{2}$Sudarshan T.S.B is a Professor at Department Of Computer Science and Engineering, PES University, Bangalore, India,
        {\tt\small sudarshan.tsb@gmail.com}}%
}

\begin{document}

\maketitle
\thispagestyle{empty}
\pagestyle{empty}

\begin{abstract}

Popular navigation stacks implemented on top of open-source frameworks such as ROS(Robot Operating System) and ROS2 represent the robot workspace using a discretized 2D occupancy grid. This method, while requiring less computation, restricts the use of such navigation stacks to wheeled robots navigating on flat surfaces.

In this paper, we present a navigation stack that uses a volumetric representation of the robot workspace, and hence can be extended to aerial and legged robots navigating through uneven terrain. Additionally, we present a new sampling-based motion planning algorithm which introduces a bi-directional approach to the Batch Informed Trees (BIT*) motion planning algorithm, whilst wrapping it with a strategy switching approach in order to reduce the initial time taken to find a path, in addition to the time taken to find the shortest path.

\end{abstract}

\section{INTRODUCTION}

Autonomous robots need to perform a multitude of tasks in order to get from one place to another without human intervention. They maintain an internal representation of the environment they operate in as well as track their pose inside the environment. They perform motion planning in order to figure out viable paths from source to destination and also use sophisticated controllers to follow the planned path while avoiding dynamic obstacles.

Navigation stacks help developers trying to build such autonomous robots by bundling together different tools and algorithms in order to provide a cohesive solution to the navigation problem. A popular navigation stack that is used widely in both research and industry is the ROS Navigation\cite{c1}, which was built on top of the open-source framework ROS\cite{c2}. Navigation2\cite{c3} or Nav2 is another stack that was proposed as an improvement to ROS Navigation, and is built on top of ROS2\cite{c4}, the second generation of the ROS framework.

While both these stacks are constantly being used and improved by the open-source robotics community, they are limited to a being able to control wheeled robots on flat surfaces. This is because both the stacks use a 2D occupancy map to represent the environment the robots operate in. Certain modifications allow developers to use the stack for planning in 2D and collision avoidance in 3D, which is usually referred to as 2.5D. However, extending the stack's to support other types of aerial and legged robots is difficult due to the base representation being a 2D map.

Graph-search and sampling-based methods are two popular techniques for path planning in robotics. The ROS2 Navigation Stack uses only the A*\cite{c10} and Dijkstra's\cite{c11} graph search algorithms in it by default. Sampling-based motion planning algorithms have been heavily used for the purpose of planning in higher dimensions due to it's probabilistic nature. One of the recent breakthroughs in sampling-based planning is the Batch Informed Trees (BIT*)\cite{c5} motion planning algorithm. It combines the batch-wise sampling feature of the Fast Marching Tree (FMT*)\cite{c6} algorithm and the elliptical heuristic aspect of the Informed RRT* algorithm\cite{c7}, resulting in a highly efficient algorithm.

However, results indicated that the RRT-Connect algorithm\cite{c12}, which is a bi-directional motion planning algorithm fared better than BIT* in most cases in terms of the time taken to find an initial path. However, RRT-Connect isn't an asymptotically optimal algorithm unlike BIT*. Also, RRT-Connect does not switch between strategies used for various modules like tree swapping, forming connections between trees, etc. The same strategy and approach is used continuously. 

In this paper, we present a navigation stack that uses a volumetric representation of the robot's environment in order to allow motion planning and control in full 3D. The proposed stack makes use of available open-source libraries, such as Octomap\cite{c13}, a 3D mapping framework that uses the Octree data structure to save volumetric data efficienty, the Open Motion Planning Library(OMPL)\cite{c14} to plan collision free paths through the map, and Robot Localization\cite{c15}, a multi-sensor fusion framework for state estimation. The stack has been tested using a custom built differential-drive robot platform. Additionally, we propose a new sampling-based motion planning algorithm, which combines features of RRT-Connect to the BIT* algorithm, whilst simultaneously introducing switching of strategies for various modules involved in the bi-directional approach.

\section{RELATED WORK}

\subsection{Navigation2}
Built on top of the industry grade, secure message-passing framework ROS2, Navigation2 is a modular software stack that provides support for autonomous navigation of holonomic, differential-drive, legged, and ackermann style robots with collision checking in SE2 space.

Navigation2 makes use of behaviour trees\cite{c16} in order to model navigation sequences and perform planning, control and recovery. This provides a simple mechanism to generate complex behaviours such as waypoint navigation. Modularity is introduced into the system by compartmentalizing individual functionalities such as state estimation, planning and control and integrating them using ROS2 servers. A similar architecture is followed in the proposed navigation stack to allow easy replacement of different components of the system, providing room for further improvement.

\subsection{Octomap}

Octomap presents a mapping framework that makes use of the Octree data structure to represent 3D space. The use of a tree data structure allows for recursive division of space into cubes known as voxels. Neighbouring voxels containing similar information can further be combined into a bigger voxel, which provides ways to compress the generated map in order to reduce memory consumption. The framework also uses a probabilistic representation of map occupancy which allows for modelling of uncertainty due to sensor noise. The use of a tree structure also allows for explicit differentiation between unknown, free and occupied space, which is important in the context of autonomous exploration.

\subsection{Motion Planning}
A bidirectional approach to the Batch Informed Trees (BIT*) algorithm has already been introduced in the form of Adaptively Informed Trees (AIT*)\cite{c8}, which is an algorithm that focuses on adaptive heuristics. A lazy reverse search takes place in the problem domain, providing a feedback to the forward search, which in turn provides feedback to the reverse search in order to improve the current heuristic. However, we are proposing an algorithm where the heuristic remains constant but the search itself is bidirectional where the trees are searching for each other with interchanging strategies depending on the situation at hand.

The paper on Context-Dependent Search for Generating Paths for Redundant
Manipulators in Cluttered Environments\cite{c9} proposes a framework for bi-directional sampling based planners for manipulators, where several strategies used for various modules of the search like  are interchanged due to scheduling. Not every strategy and module is applicable to a generic planner. Thus, we incorporate the strategies used to attempt a connection between the trees and implement a schedule for the strategy's execution on the bi-directional approach of the BIT*.

\section{DESIGN}

\subsection{System Architecture}
Fig 1 shows the overall architecture of the proposed navigation stack. Input from multiple sensors such as Inertial Measurement Unit (IMU) and wheel odometry is used to estimate the state of the robot. The estimated state information is combined with depth data from the RGBD camera to generate a 3D map. 

   \begin{figure}[thpb]
      \centering
      \framebox{\parbox{3in}{
      \includegraphics[scale=0.263]{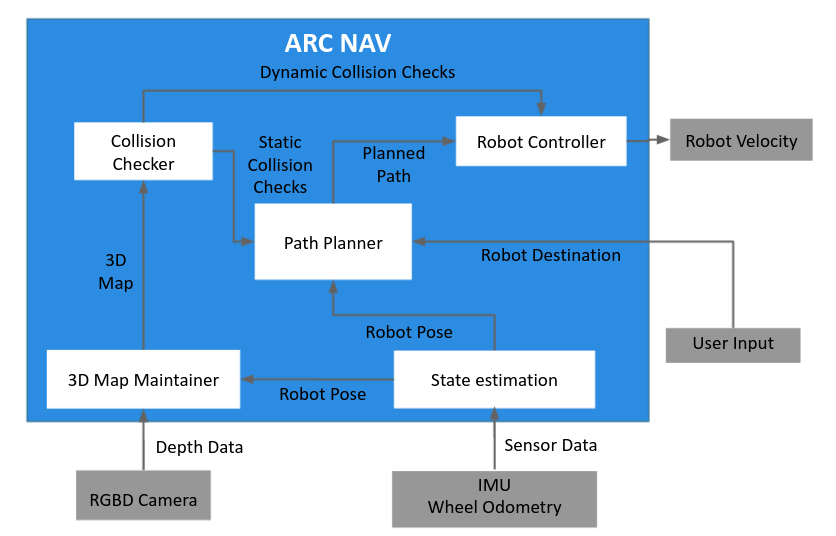}
}}
      \caption{Overview of the Navigation Stack Architecture}
      \label{figurelabel}
   \end{figure}

During navigation, the planner makes use of the generated 3D map and the robot structure to perform collision checks and generate valid trajectories from source to destination. The valid trajectory is then provided to the robot controller in order to generate the required velocity commands. 

In the proposed architecture, static obstacles are avoided during navigation due to the collision checks performed with the 3D map during the planning stage. This is based on the assumption that the static obstacles are always static and are not moved around after the creation of the map. However, remapping during navigation is possible with the proposed architecture and can handle environments with static obstacles that can be moved around, such as chairs. Dynamic obstacle avoidance is implemented as a part of the controller where the robot can wait for the obstacle to pass or actively re-plan with an updated map that marks the obstacle.

\subsection{Motion Planning Algorithm}

The sampling-based motion planning algorithm developed introduces a bidirectional approach to the Batch Informed Trees (BIT*) algorithm. Here, two trees are expanded, where the first tree is rooted at $x_{init}$ whilst the second tree is rooted at $x \epsilon X_{goal}$. Each of these trees undergoes the same procedure as the Batch Informed Trees (BIT*) algorithm with an altercation.
The goal for the first tree $\tau1$ is the actual goal state. However, the goal state for the tree$\tau2$ is the root of $\tau1$, which is the initial state of the robot. We don't check if the sampled state lies in the goal region for either of the trees. Each iteration results in the ordering of the edge list and vertex list followed by a lazy search towards the direction of the corresponding goal depending on the heuristic at hand. Here, Each tree aims on finding a connection between itself and the other tree. The heuristic keeps decreasing in size until the two trees find each other.

There are two popular strategies existing for the purpose of connecting one tree to the other. They are:-\\
1) Attempt to connect the most recently added node of one tree to the nearest node on the other tree.\\
2) Attempt to connect the most recently added node of one tree to a randomly chosen node of the other tree.\\

\begin{algorithm}

\caption{2BIT* ($x_{start} \epsilon X_{free}$, $X_{goal} \epsilon X_{free}$)}
$P\leftarrow$ Probability of Strategy 1 being chosen;\\
$1-P\leftarrow$ Probability of Strategy 2 being chosen;\\
$V1\leftarrow\{x_{start}\}$; $V2\leftarrow\{x_{goal}$\}; $E1\leftarrow\phi$; $E2\leftarrow\phi$; $\tau1\leftarrow\{V1,E1\}$; $\tau2\leftarrow\{V2,E2\}$;\\
$x_{unconn1}\leftarrow x_{goal}$; $x_{unconn2}\leftarrow x_{start}$;\\
$Q_{V1} \leftarrow \phi$; $Q_{V2} \leftarrow \phi$; $Q_{E1} \leftarrow \phi$; $Q_{E2} \leftarrow \phi$;\\
$V_{unexpnd1}\leftarrow V1$; $V_{unexpnd2}\leftarrow V2$; $x_{new1}\leftarrow x_{unconn1}$; $x_{new2}\leftarrow x_{unconn2}$;\\
$c_{i1}\leftarrow min_{v_{goal1}\epsilon x_{goal}} \{g\tau1(v_{goal1})\}$; $c_{i2}\leftarrow min_{v_{goal2}\epsilon x_{goal}} \{g\tau1(v_{goal2})\}$;\\
\While{True}{
    \If{$Q_{V1} == \phi$ and $Q_{E1} == \phi$}
    {
        $x_{reuse1}\leftarrow$ Prune($\tau1$,$x_{unconn1}$,$c_{i1}$);\\
        $x_{sampling1}\leftarrow$Sample($m$,$x_{start}$,$x_{goal}$,$c_{i1}$);\\
        $x_{new1}\leftarrow x_{reuse1}\cup x_{sampling1}$;\\ 
        $x_{unconn1}\stackrel{+}\leftarrow x_{new1}$;\\
        $Q_{V1} \leftarrow V1$;\\
    }
    \While{BestQueueValue($Q_{V1}$)$\leq$BestQueueValue($Q_{E1}$)}
    {
        ExpandNextVertex($Q_{V1}$,$Q_{E1}$,$c_{i1}$);\\
    }
    $(v_{min},x_{min})\leftarrow$ PopBestInQueue($Q_{E1}$);\\
    \eIf{$g\tau1(v_{min})+\hat{c}(v_{min},x_{min})+\hat{h}(x_{min}) < c_{i1}$}
    {
        \If{$g\tau1(v_{min})+\hat{c}(v_{min},x_{min}) < g\tau1(x_{min})$}
        {
            $c_{edge}\leftarrow c(v_{min},x_{min})$;\\
            \If{$g\tau1(v_{min})+c_{edge}+\hat{h}(x_{min}) < c_{i1}$}
            {
                \If{$g\tau1(v_{min})+c_{edge} < g\tau1(x_{min})$}
                {
                    \eIf{$x_{min}\epsilon V1$}
                    {
                        $v_{parent}\leftarrow$ Parent$(x_{min})$;\\
                        $E1\stackrel{-}\leftarrow \{(v_{parent},x_{min})\}$;\\
                    }
                    {
                        $x_{unconn1}\stackrel{-}\leftarrow\{x_{min}\}$;\\
                        $V1\stackrel{+}\leftarrow\{x_{min}\}$;\\
                        $Q_{V1}\stackrel{+}\leftarrow\{x_{min}\}$;\\
                        $V_{unexpnd1}\stackrel{+}\leftarrow\{x_{min}\}$;\\
                    }
                    $E1\stackrel{+}\leftarrow\{(v_{min},x_{min})\}$;\\
                    $c_{i1}\leftarrow min_{v_{goal1}\epsilon x_{goal}} \{g\tau1(v_{goal1})\}$
                }
            }
        }
    }
    {
        $Q_{V1}\leftarrow\phi$; $Q_{E1}\leftarrow\phi$
    }
    Expand\_Second\_Tree()
    
}
\end{algorithm}

\begin{algorithm}
\If{$Q_{V2} == \phi$ and $Q_{E2} == \phi$}
    {
        $x_{reuse2}\leftarrow$ Prune($\tau2$,$x_{unconn2}$,$c_{i2}$);
        $x_{sampling2}\leftarrow$Sample($m$,$x_{goal}$,$x_{start}$,$c_{i2}$);\\
        $x_{new2}\leftarrow x_{reuse2}\cup x_{sampling2}$;\\ 
        $x_{unconn2}\stackrel{+}\leftarrow x_{new2}$;\\
        $Q_{V2} \leftarrow V2$;\\
    }
    \While{BestQueueValue($Q_{V2}$)$\leq$BestQueueValue($Q_{E2}$)}
    {
        ExpandNextVertex($Q_{V2}$,$Q_{E2}$,$c_{i2}$);
    }    
    $(v_{min},x_{min})\leftarrow$ PopBestInQueue($Q_{E2}$);\\
    \If{$g\tau1(v_{min})+\hat{c}(v_{min},x_{min})+\hat{h}(x_{min}) < c_{i2}$}
    {
        \If{$g\tau1(v_{min})+\hat{c}(v_{min},x_{min}) < g\tau1(x_{min})$}
        {
            $c_{edge}\leftarrow c(v_{min},x_{min})$;\\
            \If{$g\tau1(v_{min})+c_{edge}+\hat{h}(x_{min}) < c_{i2}$}
            {
                \If{$g\tau1(v_{min})+c_{edge} < g\tau1(x_{min})$}
                {
                    \eIf{$x_{min}\epsilon V2$}
                    {
                        $v_{parent}\leftarrow$ Parent$(x_{min})$;\\
                        $E1\stackrel{-}\leftarrow \{(v_{parent},x_{min})\}$;\\
                    }
                    {
                        $x_{unconn2}\stackrel{-}\leftarrow\{x_{min}\}$;\\
                        $V2\stackrel{+}\leftarrow\{x_{min}\}$;\\
                        $Q_{V2}\stackrel{+}\leftarrow\{x_{min}\}$;\\
                        $V_{unexpnd2}\stackrel{+}\leftarrow\{x_{min}\}$;\\
                    }
                    $E2\stackrel{+}\leftarrow\{(v_{min},x_{min})\}$;\\
                    $c_{i2}\leftarrow min_{v_{goal2}\epsilon x_{goal}} \{g\tau1(v_{goal2})\}$
                }
            }
        }
    }
    Connect\_Trees()
\caption{Expand\_Second\_Tree}
\end{algorithm}

\begin{algorithm}
$R\leftarrow$ A random float variable between 0 and 1;\\
\eIf{$R\geq P$}
    {
        Execute Strategy 1; (See Section 3B)  
    }
    {
        Execute Strategy 2; (See Section 3B)
    }
\If{Connection between $\tau1$ and $\tau2$ is successful}
    {
        $Q_{V2} == \phi$;
        $Q_{E2} == \phi$;
        $Q_{V1} == \phi$;
        $Q_{E1} == \phi$;
    }

\caption{Connect\_Trees}
\end{algorithm}

In this paper, we will allocate a probability for the first and second strategy being chosen. Whenever the connection between trees is being attempted, a random float variable between $0$ and $1$ is generated. Depending on the value, either the first or second strategy will be chosen and the connection attempt will be made.
If the connection is successful, the vertex and edge queues of both the trees will be emptied to allow for a new batch of nodes to be sampled in addition to the existing ones. Then, the whole process repeats until the shortest path between the initial state and the goal state is found.   

\section{IMPLEMENTATION}
\subsection{Robot Platform}
The proposed stack was implemented on top of a custom built robot platform that is shown in Figure 2. The platform is run using an Intel i5 10$^{th}$ Generation Processor. An Intel Realsense D455 Depth Camera is used to generate the depth data required for mapping and obstacle detection. The robot contains an on-board MPU9250 IMU and wheel encoders for sensor feedback.

   \begin{figure}[thpb]
      \centering
      \framebox{\parbox{2.59in}{
      \includegraphics[scale=0.062]{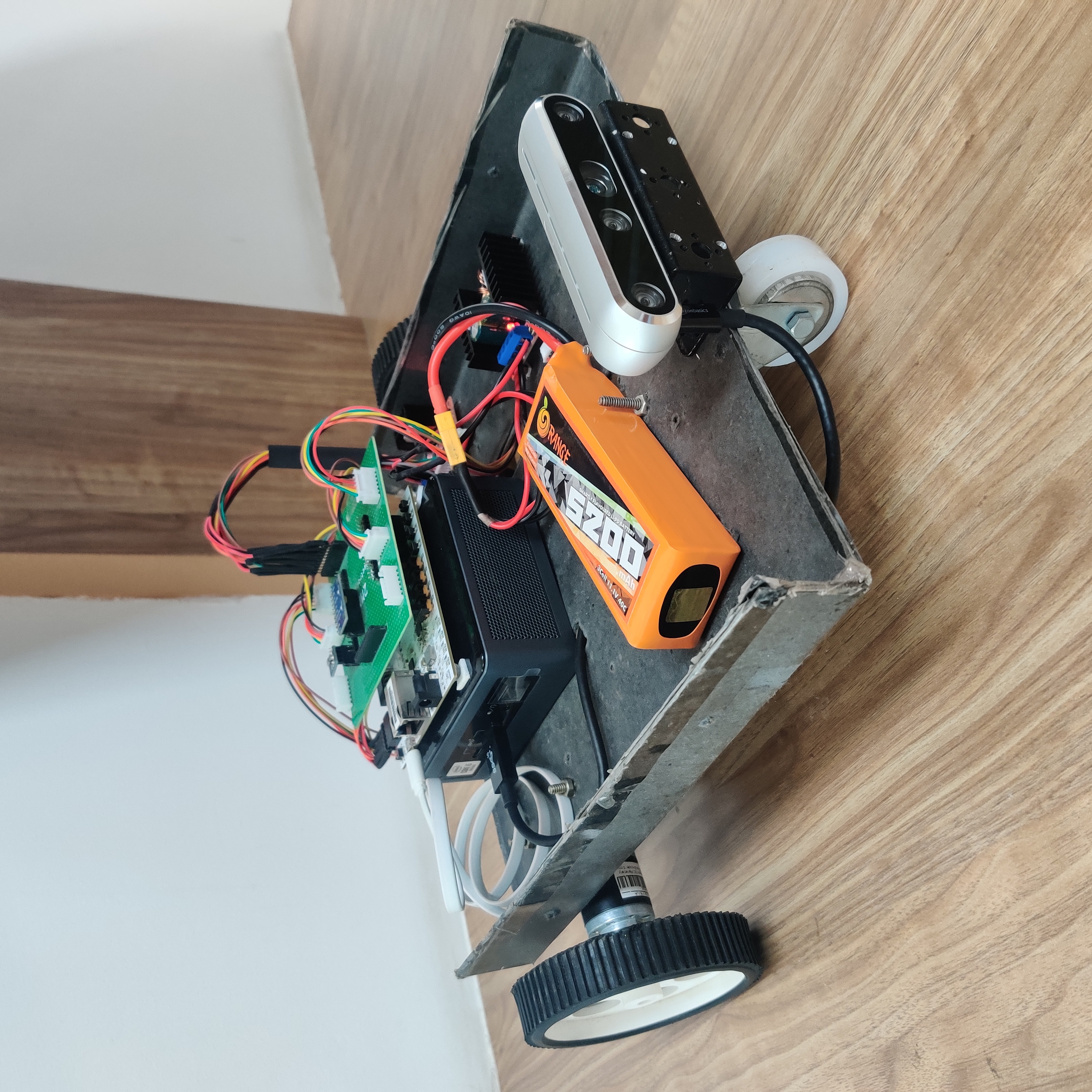}
}}
      \caption{Robot Platform used to test the Navigation Stack}
      \label{figurelabel}
   \end{figure}

\subsection{State Estimation}

Sensor feedback from the on-board IMU and odometery data from the wheel encoders are used to estimate the pose of the robot in real-time. This is implemented using Robot Localisation, which uses an Extended Kalman Filter to fuse the sensor data and obtain a reliable estimation.

\subsection{Mapping}

Pose data generated through state estimation is used to combine different depth data frames in order to generate a 3D map of the robot workspace. This is implemented using Octomap, which provides functionalities to insert depth data at a particular pose, serialize map data and save it as a file, and to retrieve the map for later use.

\subsection{Planning}
The generated 3D map is used to perform motion planning through OMPL. The planning problem is constructed as a graph-search with a node in the graph representing the state of the robot, which in this case is it's pose. RRT-Connect, a random sampling-based algorithm is used to perform the graph search, and the Flexible Collision Library (FCL)\cite{c16} is used to perform collision checks between the robot mesh and the 3D map to identify valid states.

\section{RESULTS}

Figure 3 shows a comparison between the 3D map generated using the proposed navigation stack and a portion of the lab. Figure 4 shows a 3D map of a corridor inside the campus. Figure 5 shows the top view of a path planned through the 3D map of the lab using OMPL. The green line in Figure 5 denotes the collision-free path from the robot's position to a user-defined destination. 

Figure 6 shows a 2D map of the same lab, generated using a cross section of the 3D map at the height of the sensor, which is similar to a typical 2D map generated using a laser scanner. There is a high chance of important features of the environment including the corners of table tops and chairs, spikes on irregular surfaces, etc. being missed by the 2D map generated. This can lead to generation of unsafe paths during the planning stage, which further leads to collisions during the control stage.

   \begin{figure}[thpb]
      \centering
      \framebox{\parbox{3in}{
      \includegraphics[scale=0.195]{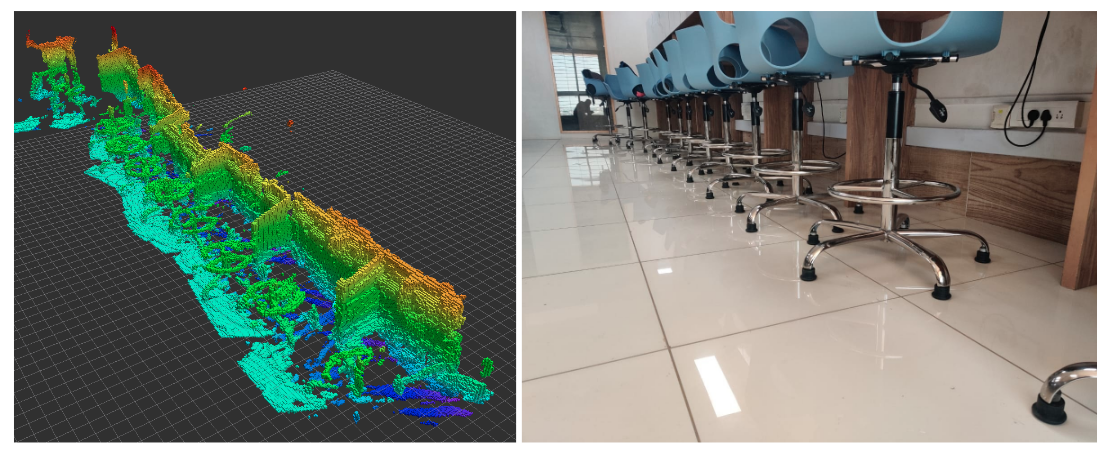}
}}
      \caption{3D Map Comparison}
      \label{figurelabel}
   \end{figure}
   
   \begin{figure}[thpb]
      \centering
      \framebox{\parbox{3in}{
      \includegraphics[scale=0.26]{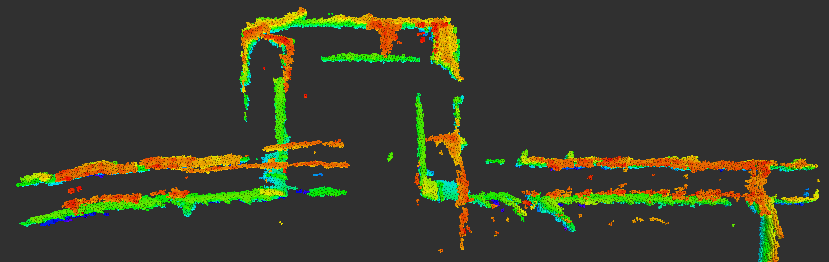}
}}
      \caption{3D Map of Campus Corridor}
      \label{figurelabel}
   \end{figure}

   \begin{figure}[thpb]
      \centering
      \framebox{\parbox{3in}{
      \includegraphics[scale=0.5]{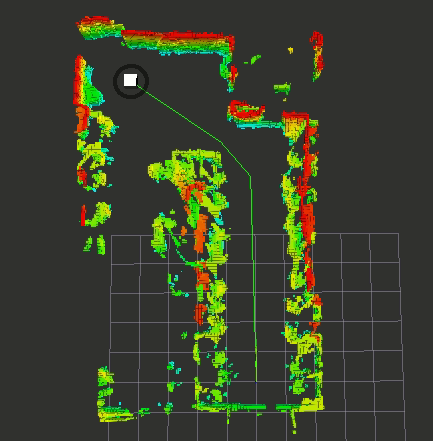}
}}
      \caption{Motion Planning through 3D map of the lab.}
      \label{figurelabel}
   \end{figure}
   
   \begin{figure}[thpb]
      \centering
      \framebox{\parbox{3in}{
      \includegraphics[scale=0.59]{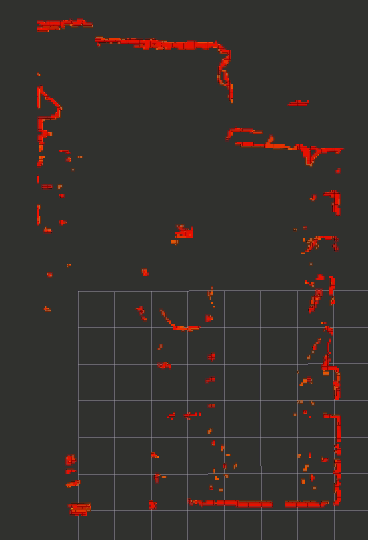}
}}
      \caption{2D map of the lab at sensor height}
      \label{figurelabel}
   \end{figure}

\section{CONCLUSION AND FUTURE SCOPE}

In this paper, we have proposed and tested a 3D navigation stack that can perform state estimation, mapping, motion planning and control. Performing these operations in 3D allows us to use the same system for navigation on legged and aerial robots, which require a full representation of the environment. Navigation on such platforms will be tested and improved. We have also proposed a sampling based motion planning algorithm which introduces a bidirectional approach to the BIT* algorithm. 
The future scope of this project is to implement the proposed sampling based algorithm and perform an in depth analysis and benchmarking on it. Additionally, we plan on appending this algorithm to the Open Motion Planning Library (OMPL), which would be connected to this navigation stack indirectly.

\addtolength{\textheight}{-12cm}   




\section*{ACKNOWLEDGMENT}

We would like to thank Dr. Shikha Tripathi, Head of Robotics, Automation and Intelligent Systems (RAIS) lab, PES University, for providing us access to the lab space and resources.


\end{document}